\documentclass{article}

\usepackage{arxiv}

\usepackage[utf8]{inputenc} 
\usepackage[T1]{fontenc}    
\usepackage[hidelinks,breaklinks=true]{hyperref}       
\usepackage{url}            
\usepackage{booktabs}       
\usepackage{amsfonts}       
\usepackage{nicefrac}       
\usepackage{microtype}      
\usepackage{graphicx}
\usepackage{natbib}
\usepackage{doi}
\usepackage{algorithm}
\usepackage{algpseudocode}
\usepackage{multirow}
\usepackage{mathtools}
\usepackage{breakcites}

\usepackage[acronym]{glossaries}
\makeglossaries

\newacronym{ai}{AI}{Artificial Intelligence}
\newacronym{ml}{ML}{Machine Learning}
\newacronym{rl}{RL}{Reinforcement Learning}
\newacronym{vae}{VAE}{Variational Autoencoder}
\newacronym{gce}{GCE}{Graph Counterfactual Explainability}
\newacronym{gnn}{GNN}{Graph Neural Network}
\newacronym{cf2}{CF$^2$}{Counterfactual and Factual}
\newacronym{meg}{MEG}{Molecular Explanation Generator}
\newacronym{clear}{CLEAR}{CounterfactuaL ExplAnation geneRator for graphs}
\newacronym{gdpr}{GDPR}{General Data Protection Regulation}
\newacronym{our}{DyGRACE}{\textbf{Dy}namic \textbf{GRA}ph \textbf{C}ounterfactual \textbf{E}xplainer}
\newacronym{gae}{GAE}{graph autoencoder}
\newacronym{ged}{GED}{Graph Edit Distance}

\title{Adapting to Change: Robust Counterfactual Explanations in Dynamic Data Landscapes}

\date{} 					

\author{Bardh Prenkaj\inst{1} \and Mario Villaizán-Vallelado\inst{2,}\inst{3} \and Tobias Leemann\inst{4} \and Gjergji Kasneci\inst{5}} 

\author{ \href{https://orcid.org/0000-0002-2991-2279}{\includegraphics[scale=0.06]{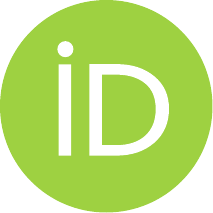}\hspace{1mm}Bardh Prenkaj}\\
	Department of Computer Science\\
	Sapienza University of Rome\\
	Rome, Italy \\
	\texttt{prenkaj@di.uniroma1.it}
	\And
	\href{https://orcid.org/0009-0002-0754-1742}{\includegraphics[scale=0.06]{orcid.pdf}\hspace{1mm}Mario Villaizán-Vallelado} \\
	University of Valladolid\\
    Artificial Intelligence Laboratory (AI-Lab), Telefónica I+D \\
	Valladolid, Spain \\
	\texttt{mario.villaizan@uva.es}\\
 \texttt{mario.villaizanvallelado@telefonica.com} \\
    \And
    \href{https://orcid.org/0000-0001-9333-228X}{\includegraphics[scale=0.06]{orcid.pdf}\hspace{1mm}Tobias Leemann} \\
	University of Tuebingen\\
	Tuebingen, Germany \\
	\texttt{tobias.leemann@uni.tuebingen.de}
    \And
    \href{https://orcid.org/0000-0002-3123-7268}{\includegraphics[scale=0.06]{orcid.pdf}\hspace{1mm}Gjergji Kasneci} \\
	TU Munich\\
	Munich, Germany \\
	\texttt{gjergji.kasneci@tum.de} 
}

\renewcommand{\shorttitle}{Adapting Counterfactual Explanations in Dynamic Data Landscapes}

\hypersetup{
pdftitle={Adapting to Change: Robust Counterfactual Explanations in Dynamic Data Landscapes},
pdfsubject={explainability},
pdfauthor={Bardh Prenkaj,Mario Villaizan-Vallelado,Tobias Leemann,Gjergji Kasneci},
pdfkeywords={Graph Counterfactual Explainability, Dynamic Explainability, Time-dependent Explanations},
}

\begin{document}
\maketitle

\begin{abstract}
	We introduce a novel semi-supervised Graph Counterfactual Explainer (GCE) methodology, Dynamic GRAph Counterfactual Explainer (\acrshort{our}). It leverages initial knowledge about the data distribution to search for valid counterfactuals while avoiding using information from potentially outdated decision functions in subsequent time steps. Employing two graph autoencoders (GAEs), \acrshort{our} learns the representation of each class in a binary classification scenario. The GAEs minimise the reconstruction error between the original graph and its learned representation during training. The method involves (i) optimising a parametric density function (implemented as a logistic regression function) to identify counterfactuals by maximising the factual autoencoder's reconstruction error, (ii) minimising the counterfactual autoencoder's error, and (iii) maximising the similarity between the factual and counterfactual graphs. This semi-supervised approach is independent of an underlying black-box oracle. A logistic regression model is trained on a set of graph pairs to learn weights that aid in finding counterfactuals. At inference, for each unseen graph, the logistic regressor identifies the best counterfactual candidate using these learned weights, while the GAEs can be iteratively updated to represent the continual adaptation of the learned graph representation over iterations. \acrshort{our} is quite effective and can act as a drift detector, identifying distributional drift based on differences in reconstruction errors between iterations. It avoids reliance on the oracle's predictions in successive iterations, thereby increasing the efficiency of counterfactual discovery. \acrshort{our}, with its capacity for contrastive learning and drift detection, will offer new avenues for semi-supervised learning and explanation generation.
\end{abstract}


\section{Introduction}\label{sec:intro}

In the era of big data and complex \acrlong{ml} models, explainability and interpretability have emerged as critical aspects, not only for the technical merits of transparency and robustness but also from a regulatory perspective. With regulations such as the European Union's \acrfull{gdpr} and the proposed \acrlong{ai} Act, there is a growing demand for models that perform well and provide interpretable and actionable insights into their predictions. A key element of meeting these regulatory demands in a human-centred way is using counterfactual explanations, which illuminate model decisions by illustrating alternative scenarios that would lead to different outcomes.

However, as shown by recent work \cite{pawelczyk2023trade}, a significant challenge arises when we consider the dynamic and ever-evolving nature of the data these models interact with. Data undergoes continuous changes and distribution shifts, which can critically impact the robustness, relevance, and, therefore, the validity of counterfactual explanations. Existing solutions have yet to adequately address this complex interplay between robust counterfactual generation and dynamic data landscapes.

We dive into the challenge of generating robust counterfactual explanations under data changes and distribution shifts by providing a novel technique for representing and tracking data throughout temporal changes. This under-researched area is increasingly important, as meeting compliance needs in rapidly evolving real-world scenarios is paramount. Our research outlines and empirically evaluates a novel approach that adaptively generates robust counterfactual explanations, which meets the demands of model transparency and understanding and provides a basis for current regulatory requirements. This work aims to contribute to the discourse on \acrshort{ai} interpretability, ethics, and regulation, helping to create \acrlong{ml} models that remain transparent, accountable, and compliant, even amidst changing data landscapes.

\section{Related Work}\label{sec:related_work}

To the best of our knowledge, this is the first work on \acrfull{gce} considering distributional drift happening in time. While updating (or even retraining) the prediction model under distributional drifts has been extensively explored \cite{bashir2017framework, haug2021learning, lughofer2016recognizing, sethi2017reliable}, aligning counterfactual explanations after a drift happens is yet to be covered. Only Pawelczyk et al. \cite{pawelczyk2023trade} tackle the problem of \textit{recourse} (i.e., counterfactual) fragility when data is deleted in the future. The authors pinpoint the most influential data points such that their deletion at time $t+\delta$ ensures the obsoleteness of generated counterfactuals at a previous time $t$. Here, we propose a semi-supervised explanation approach that produces counterfactuals in a data-driven and principled way and integrates a drift detection mechanism to signal counterfactual invalidity, thus updating the explainer to produce valid counterfactuals again.

We provide the reader with the most recent time-unaware \acrshort{gce} approaches for completeness. Recently, time-unaware \acrshort{gce} has received more attention due to the upsurging phenomenon of the need for explainability in graph domains such as fraud detection in bank transactions \cite{dumitrescu2022}, drug-disease comorbidity prediction \cite{madeddu2020feature}, and community detection \cite{wu2022clare}. Prado-Romero et al. \cite{prado2022survey} provide a thorough survey on \acrshort{gce} and categorise the methods according to three classes: i.e., search-, heuristic-, and learning-based approaches. 

\noindent\textbf{Search- and heuristic-based approaches} rely on a specific criterion, such as the similarity between instances, to search for a suitable counterfactual within the dataset. Contrarily, heuristic-based methods adopt a systematic approach to modify the input graph until a valid counterfactual is obtained.

\noindent \textit{DCE} \cite{faber2020contrastive} aims to find a counterfactual graph $G'$ similar to the input graph $G$ but belonging to a different class. In the realm of graph counterfactuality \cite{abrate2021counterfactual}, \textit{DDBS} and \textit{OBS} are two heuristic approaches used in brain networks. These methods represent the brain as a graph with vertices denoting regions of interest (ROIs) and edges representing connections between co-activated ROIs. Both DDBS and OBS employ a bidirectional search heuristic. Initially, they perturb the edges of the input graph $G$ until a counterfactual graph $G'$ is achieved. Subsequently, they refine the perturbations to reduce the distance between $G$ and $G'$ while ensuring the counterfactual condition. 

\noindent \textit{RCExplainer} \cite{bajaj2021robust} utilizes a \acrshort{gnn} to define decision regions with linear boundaries, capturing shared characteristics of instances within each class. Unsupervised methods identify these regions, preventing overfitting due to potential noise or peculiarities in specific instances. A loss function based on these boundaries then trains a network to select a small subset of edges $E^*$ from the original graph $G$. The resulting graph $G^* = (V^*, E^*)$, belonging to the same class as $G$, can be transformed into a counterfactual graph $G'$ outside this decision region, satisfying the counterfactual condition.

\noindent We point the reader to \cite{huang2023global,liu2021multi,wellawatte2022model} for other search- and heuristic-based methods.

\noindent\textbf{Learning-based approaches} share a three-step pipeline: 1) generating masks that indicate the relevant features given a specific input graph $G$; 2) combining the mask with $G$ to derive a new graph $G'$; 3) feeding $G'$ to the prediction model (oracle) $\Phi$ and updating the mask based on the outcome $\Phi(G')$. Generally, learning-based strategies for generating counterfactual explanations can be categorised into three main groups: i.e., perturbation matrix \cite{cai2022probability, lucic2022cf, tancf2, wellawatte2022model, wu2021counterfactual}, \acrlong{rl} (\acrshort{rl}) \cite{nguyen2022explaining, numeroso2021meg}, and generative approaches \cite{ma2022clear,sun2021preserve}. Here, we describe the most interesting for each category.

\noindent\textit{\acrshort{cf2}} \cite{tancf2} produces factual explanations by balancing factual and counterfactual reasoning. It generates a factual subgraph, a subset of the original input graph, and then derives a counterfactual by removing this factual subgraph, following a similar approach as described in \cite{bajaj2021robust}.

\noindent\textit{\acrshort{meg}} \cite{numeroso2021meg} and \textit{MACCS} \cite{wellawatte2022model} use multi-objective \acrshort{rl} to generate molecule counterfactuals. However, their domain-specificity limits their applicability to other domains. The reward function includes a task-specific regularisation term to guide perturbation actions. Similarly, MACDA \cite{nguyen2022explaining} employs \acrshort{rl} for counterfactual generation in drug-target affinity prediction. 

\noindent\textit{\acrshort{clear}} \cite{ma2022clear} is a generative method that utilises a \acrfull{vae} to generate counterfactuals. The counterfactuals produced are complete graphs with stochastic edge weights. To obtain valid counterfactuals, a sampling procedure is employed. Graph matching between $G$ and $G'$ is required due to potential differences in vertex order, which can be time-consuming \cite{livi2013graph}.

\noindent A recent paper on generative approaches for \acrshort{gce} \cite{prado2023revisiting} explored the adaptation of \textit{CounteRGAN} \cite{nemirovsky22} in the graph domain. The authors show how generative strategies are useful to generate multiple counterfactuals without relying on the oracle at inference time. However, these approaches need to be further explored since they do not reach satisfactory performances.

\section{Problem Formulation}\label{sec:problem}

We consider prediction problems $\Phi \colon G \rightarrow Y$ where $G = \left( V, E \right)$ is a graph with vertex and edge sets $V = \left\lbrace v_1, \dots, v_n \right\rbrace$ and $E = \left\lbrace \left( v_i, v_j \right) \; \middle| \; v_i, v_j \in V \right\rbrace$, respectively, and $Y$ is the set of classes; w.l.o.g., we assume $Y \in \left\lbrace 0, 1 \right\rbrace$. We denote with $\mathcal{G} = \left\lbrace G_1, \dots, G_k \right\rbrace$ the dataset containing different graphs $G_i \; \forall i \in \left[ 1, k \right]$. According to Prado-Romero et al. \cite{prado2022survey}, the ``closest'' counterfactual $\mathcal{E}_\Phi \left( G_i \right)$ of $G_i$, given the classifier (oracle) $\Phi$, can be defined as in Eq. \ref{eq:binary_counterfactual}.
\begin{equation}\label{eq:binary_counterfactual}
    \mathcal{E}_{\Phi} \left( G_i \right) = \underset{G_j^{\prime} \in \mathcal{G}^{\prime}, G_i \neq G_j^{\prime}, \Phi \left( G_i \right) \neq \Phi \left( G_j{\prime} \right)}{\arg\max} \mathcal{S} \left( G_i, G_j^{\prime} \right)
\end{equation}
where $\mathcal{G}^{\prime}$ is the set of all possible graphs, and $\mathcal{S} \left( G_i, G_j^{\prime} \right)$ measures the similarity between the graph $G_i$ and its counterfactual $G^{\prime}_j$. Notice that Eq. \ref{eq:binary_counterfactual} produces a single counterfactual\footnote{In case multiple counterfactuals maximise this probability, one can break ties arbitrarily to produce a single one.} instance $G_j^{\prime}$ that is the most similar to $G_i$. The \textit{search} for the counterfactuals is conditioned such that the returned instance $G_j^{\prime}$ is different\footnote{Some methods \cite{abrate2021counterfactual} default to the original instance if the search/heuristic fails to produce a valid counterfactual.} from the original $G_i$. 

Although Eq. \ref{eq:binary_counterfactual} has been widely adopted in the literature, the usage of the similarity metric to produce counterfactuals is loosely defined because different metrics might produce different counterfactuals for the same input graph $G_i$. To this end, we take a probabilistic perspective and aim to generate a counterfactual instance that is quite likely within the distribution of valid counterfactuals by maximising Eq. \ref{eq:stochastic_counterfactuality}. 

\begin{equation}\label{eq:stochastic_counterfactuality}
    \mathcal{E}_{\Phi} \left( G_i \right) = \underset{G_j \in \mathcal{G}}{\arg \max} \; P_{cf} \left( G_j, \; \middle| \; G_i , \Phi\left(G_i\right) , \neg \Phi \left( G_i \right)\right)
\end{equation}
Here, we use the notation $\neg \Phi \left( G_i \right)$ to indicate any other class from the one predicted for $G_i$, thus supporting also multi-class classification problems. In a binary classification scenario, $\neg \Phi \left( G_i \right)$ becomes $1-\Phi \left( G_i \right)$.

\begin{figure}[!t]
    \centering
    \includegraphics[width=\textwidth]{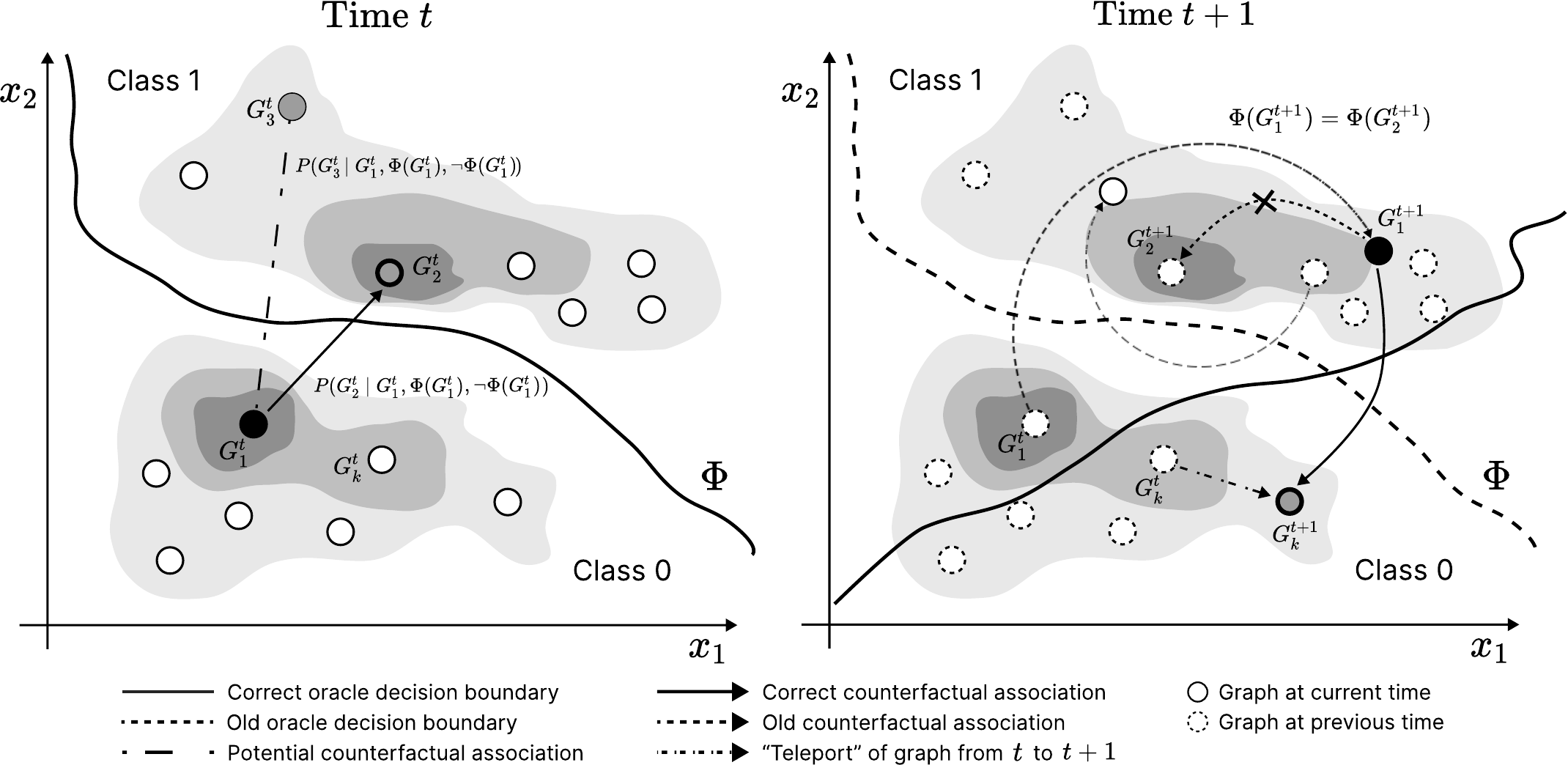}
    \caption{Counterfactuality under distributional drifts. (left) Given the decision boundary of the oracle $\Phi$ trained on the data at time $t$, graph $G_2^t$ is correctly associated as the counterfactual of $G_1^t$ since it satisfies Eq. \ref{eq:timed_stochastic_counterfactuality}. (right) Drift happens and $G^t_1$ is "teleported" at $G_1^{t+1}$ crossing the old (dotted red line) decision boundary. Here, $G_2^{t+1}$ cannot be a counterfactual for $G_1^{t+1}$ since $\Phi \left( G_1^{t+1} \right) = \Phi \left( G_2^{t+1} \right)$. Assuming that $G_k^{t+1}$ satisfies Eq. \ref{eq:timed_stochastic_counterfactuality} at time $t+1$, we can signal a drift and potentially update $\Phi$'s decision boundary (green full line), thus changing the counterfactuals where applicable.}
    \label{fig:timed_counterfactuality}
\end{figure}

As anticipated in Sec. \ref{sec:related_work}, counterfactual validity is defied when distributional drifts happen in time. Now, for different time stamps, we have different snapshots of the same dataset, $\mathcal{G}^t = \left\lbrace G_1^t, \dots, G_k^t \right\rbrace$ where $t \in \left[ 0, T \right]$ and $T$ is the maximum monitoring time. At any particular time $t+1$, it might happen that the oracle wrongly predicts the class for $G_i^t$, i.e., $\Phi \left( G_i^t \right) \neq \Phi \left( G_i^{t+1} \right)$. This means that $G_i^t$ has experienced changes in its structure, which led to a change of its class at time $t+1$. If $\Phi \left( G^t_i \right) \neq \Phi \left( G^{t+1}_i \right)$, we expect that the counterfactual for $G_i^{t+1}$ to change w.r.t. that of $G^t_i$. Therefore, we take into account the time factor to redefine Eq. \ref{eq:stochastic_counterfactuality} as follows:
\begin{equation}\label{eq:timed_stochastic_counterfactuality}
    \mathcal{E}_{\Phi}\left( G_i^t \right) = \underset{G_j^t \in \mathcal{G}}{\arg \max}\; P_{cf}^t \left( G_j^t,  \; \middle| \; G_i^t, \Phi \left( G_i^t \right), \neg \Phi\left( G_i^t \right) \right)
\end{equation}
To the best of our knowledge, this is the first work that tries to integrate drift detection with counterfactuality change in time. In other words, we can signal a drift happening if $\mathcal{E}_{\Phi} \left( G_i^t \right) \neq \mathcal{E}_{\Phi} \left( G_i^{t+1} \right)$ because it means that the original graph $G_i^t$ has moved beyond the decision boundary of $\Phi$ at time $t+1$ (see Fig. \ref{fig:timed_counterfactuality}). In these scenarios, we can trigger an update of $\Phi$ to reflect the changes after the drift and regenerate the counterfactuals accordingly. However, in cases where $G_i$ has changed from $t$ to $t+1$ but has $\Phi \left( G_i^t \right) = \Phi \left( G_i^{t+1} \right)$, then, even though its counterfactual might change structure (see Eq. \ref{eq:timed_stochastic_counterfactuality}), it still remains valid, maintaining the opposite class. Here, a full update of $\Phi$ could be avoided in real-world scenarios.

\section{Methodology}\label{sec:method}

Here, we describe our method, \acrlong{our}, namely \acrshort{our}\footnote{We provide our implementation in \url{https://github.com/bardhprenkaj/HANSEL}.}. \acrshort{our} is a semi-supervised \acrshort{gce} method that uses $\Phi$ in the first time step $t_0$ to obtain knowledge about the data distribution and search for valid counterfactuals while avoiding getting hints from its (possibly) outdated decision function for $t_i > t_0 \; \forall i \in \left[ 1, T \right]$. To favour readability, we omit the time superscript from the formulas unless necessary for disambiguation.

Recall that we are in a binary classification scenario. However, the following observations can be easily extended to a multi-class classification problem. Here, we rely on two \acrlong{gae}s (\acrshort{gae}s) \cite{kipf2016variational}, i.e., $f_y, f_{\neg y}: \mathcal{G} \rightarrow \mathcal{G}$ that are responsible for learning how to represent each class in $Y$, respectively. At each time step, $G_i \in \mathcal{G}$ gets directed through one of the autoencoders based on $y = \Phi \left( G_i \right)$. The objective of each autoencoder during the training phase is to minimise the reconstruction error between the original graph $G_i$ and its learned representation $\hat{G}_i$. Generally, the reconstruction score is a function $h: \mathcal{G} \times \mathcal{G} \rightarrow \mathbb{R}$. For a pair of instances $\left( G_i, G_j \right)$ s.t. $\Phi \left( G_i \right) \neq \Phi \left( G_j \right)$, we expect $h(G_j, f_{y}\left( G_j \right)) \geq h(G_j, f_{\neg y}\left( G_j \right))$. This is the case since the autoencoder corresponding to the counterfactual class should know how to represent $G_j$, thus having a low reconstruction error. Contrarily, the autoencoder corresponding to the factual class should not be able to (at least not easily) reconstruct a counterfactual. 

Once $f_y$ and $f_{\neg y}$ are trained, \acrshort{our} maximises the probability in Eq. \ref{eq:timed_stochastic_counterfactuality} to find counterfactuals for all $G_i \in \mathcal{G}$. We model $P_{cf}$ by the parametric density function in Eq. \ref{eq:optimisation_func} where $\alpha$, $\beta$, and $\gamma$ are learned weights.
\begin{equation}\label{eq:optimisation_func}
    \begin{gathered}
        \underset{G_j \in \mathcal{G}}{\arg\max}\; P_{cf} \left( G_j \; \middle| \; G_i, \Phi \left( G_i \right), \neg \Phi \left( G_i \right) \right) \\
        = \underset{G_j \in \mathcal{G}}{\arg\max} \left( \alpha h(G_j, f_{y} \left( G_j \right)) - \beta h(G_j, f_{\neg y} \left( G_j \right)) + \gamma g \left( G_i, G_j \right) \right)
    \end{gathered}
\end{equation}
where $y = \Phi \left( G_i \right)$, $g: \mathcal{G} \times \mathcal{G} \rightarrow \mathbb{R}$ measures the similarity between two graphs, and $\alpha$, $\beta$, and $\gamma$ are learned weights. Eq. \ref{eq:optimisation_func} maximises the reconstruction error of $G_j$ from the factual autoencoder, minimise - hence the negation - the error of $G_j$ from the counterfactual autoencoder, and maximises the similarity of $G_j$ w.r.t. $G_i$. In other words, we search for counterfactuals that are far away\footnote{One can see the similarity function $g$ as the specular of a particular distance function, provided that this has a codomain of $\mathbb{R}_{0}^{1}$.} from other factual graphs besides $G_i$.

Notice that the weights $\alpha$, $\beta$, and $\gamma$ in Eq. \ref{eq:optimisation_func} can be solved via a logistic regression trained on a set of graph pairs. We assign pairs of graphs with $\left( G_i, G_j \right)$ a label of $1$ if $\Phi \left( G_j \right) \neq \Phi \left( G_i \right)$, and $0$ otherwise. By solving this objective function, we can interpret the learned weights and assess the contribution of each component in the equation of finding the counterfactuals for each $G_i$. At inference time, we get a never-seen before graph $G^*$, and for all $G_i \in \mathcal{G}$, we calculate the reconstruction errors $h(G_i, f_{0}\left( G_i \right))$ and $h(G_i, f_{1} \left( G_i \right))$, and the similarity $g \left( G^*, G_i \right)$.  We use these values as input to the trained logistic regressor to find the ``best'' counterfactual candidate for $G^*$.

In the next iterations, we do not rely on $\Phi$'s predictions since they might not represent the reality of the new incoming data (see Fig. \ref{fig:timed_counterfactuality}). Instead, we exploit the learned representation of the two \acrshort{gae}s from the previous iteration. In other words, for each $G_i^t \in \mathcal{G}$ s.t. $t \in [1,T]$, we exploit the reconstruction errors $h(G_i^t, f_0(G_i^t))$ and $h(G_i^t, f_1(G_i^t))$ to find the label of $G_i^t$. Notice that one of the \acrshort{gae}s embodies the latent representation of $G_i^t$, thus producing a smaller reconstruction error and playing the role of the factual autoencoder. Now, we can use the logistic regressor trained in at time $t-1$ to return the counterfactual $G_j^t \in \mathcal{G}$. 

In practice, to support a continual adaptation of the learned graph representation of the \acrshort{gae}s, we find the top $k$ counterfactuals via the logistic regressor. We use these instances to update the knowledge of the counterfactual \acrshort{gae} and minimise the reconstruction error. Contrarily, we can use the same counterfactual candidates to maximise their reconstruction error by the factual \acrshort{gae}, thus steering it away from the counterfactual representation space. This goes in hand with the intuition of contrastive learning since the factual \acrshort{gae}, at each iteration, learns to be specific about the factual instances and is drawn away from potential counterfactuals. The same reasoning applies to the counterfactual \acrshort{gae}. After each iteration, the logistic regressor can be updated (or even trained from scratch) on the ``newly gained'' knowledge of the \acrshort{gae}s. In this way, the prediction decision function gets mimicked by the autoencoders instead of an external (possibly) black-box oracle $\Phi$.

Recall that we do not rely on the oracle $\Phi$ predictions in successive iterations but on the learned representation of the \acrshort{gae}s at previous ones. Nevertheless, \acrshort{our} can play the role of a drift detector based on the reconstruction errors at iteration $t$ and those at $t-1$. In other words, $f_y$ and $f_{\neg y}$ can be used to measure the reconstruction errors for $G_i^{t-1} \in \mathcal{G}$ based on $y = \Phi \left( G_i^{t-1} \right)$. Then, we can do the same for $G_i^t \in \mathcal{G}$ based on $y = \Phi \left( G_i^{t} \right)$. If the distributions of the reconstruction errors at $t$ and $t-1$ are different according to a statistic test (e.g., Kolmogorov-Smirnov test), then we can signal a distributional drift and update $\Phi$ accordingly. Afterwards, $f_y$ and $f_{\neg y}$ are retrained according to the updated $\Phi$, and Eq. \ref{eq:optimisation_func} is optimised. However, notice that this procedure is supervised and depends on $\Phi$, which does not guarantee satisfactory performances at each iteration to guide the search for valid counterfactuals. Exploiting the learned representation of the \acrshort{gae}s in a semi-supervised manner as described above is more efficient and decouples itself from the underlying oracle $\Phi$.

\section{\acrshort{our}'s performance analysis}\label{sec:experiments}
Here, we assess the performances of \acrshort{our} and the other SoTA methods. First, we describe the adopted benchmarking datasets providing the details on their generation process (see Sec. \ref{sec:datasets}). Then, we describe the evaluation metrics and the hyperparameters used to run each method (see Sec. \ref{sec:metrics_and_hyperparams}). Finally, in Sec. \ref{sec:discussion}, we provide a discussion of the performance of \acrshort{our}.

\subsection{Benchmarking Datasets}\label{sec:datasets}

\begin{table}[!t]
\centering
\caption{The dataset characteristics. $|\mathcal{G}|$ is the number of instances; $\mu(|V|)$ and $\sigma(|V|)$ represent the mean and std of the number of vertices per instance; $\mu(|E|)$ and $\sigma(|E|)$ represent the mean and std of the number of edges per instance; $|C_i|$ is the number of instances in class $i \in \{0,1\}$. $|T|$ represents the number of snapshots.}
\resizebox{.8\textwidth}{!}{\begin{tabular}{lcccccccccc}
    \toprule
     &   $|T|$ & $|\mathcal{G}|$ & &  $\mu(|V|) \pm \sigma(|V|)$ & & $\mu(|E|) \pm \sigma(|E|)$ & & $|C_0|$ & & $|C_1|$\\
    \midrule
    DynTree-Cycles & 4 & 100 & & 28 $\pm$ 0.00 & & 27.62 $\pm$ 0.645 & & 45.75 & & 54.25\\
    DBLP-Coauthors & 10 & 36 & & 13 $\pm$ 0.00 & & 41.26 $\pm$ 6.69 & & 27.27 & & 8.73\\
    \bottomrule
        \end{tabular}}
 
\label{tab:gr_datasets}
\end{table}
We test \acrshort{our} on a synthetic dataset, namely Tree-Cycles, generated according to \cite{prado2022gretel,ying2019gnnexplainer}, and a real dataset, namely DBLP-Coauthors \cite{benson2018simplicial}. See Table \ref{tab:gr_datasets} for the dataset characteristics averaged over the different time snapshots.

Tree-Cycles \cite{ying2019gnnexplainer} contains cyclic (1) and acyclic (0) graphs. We extend this dataset by introducing the time dimension, allowing graphs to evolve while maintaining class membership. We repeat the dataset generation in \cite{prado2022survey} at each time step. In this way, a particular graph $G_i^t$ can change its structure in $t+1$ and remain in the same class or move to the opposite one. This emulates a synthetic process of tracing the evolution of the graphs in the dataset according to time. Here, we guarantee that the number of instances per snapshot is the same.

The DBLP-Coauthors dataset comprises graphs representing authors, where edges denote co-authorship relationships, and edge weights signify the number of collaborations in a given year. We focus on the time frame [2000, 2010] and consider ego-networks of authors with at least ten collaborations in 2000. From this set, we randomly sample 1\% due to the dataset's scale. To trace the ego-network evolution from 2000 to 2010, we propagate ego-networks from the previous year whenever an author has no collaborations in a specific year $t$. Ego-networks are labelled 1 if their mean sum of edge weights is in the 75th percentile of average collaborations for a particular year $t$, otherwise labelled as 0.

\subsection{Evaluation metrics and hyperparameter choice}\label{sec:metrics_and_hyperparams}

We follow the suggestion in \cite{prado2022survey} to evaluate each method and use multiple metrics to show a complete and fair assessment. To this end, we exploit \textit{Runtime}, \textit{Oracle calls} \cite{abrate2021counterfactual},  \textit{Correctness} \cite{guidotti2022counterfactual,prado2022gretel}, \textit{Sparsity} \cite{prado2022gretel,yuan2022explainability}, and \textit{\acrlong{ged}} \cite{prado2022survey} as evaluation metrics. Since we return a list of counterfactuals for each input graph, we evaluate \acrshort{our} by reporting values of the previous metrics $@1,\dots,@k$.

Notice that \acrshort{our} is a flexible framework which can take any encoder-decoder combination to learn meaningful graph representations. Here, we rely on a 2-layer GCN encoder interleaved with ReLU activation functions. The output dimension of each convolution operation is 8. The decoder is a simple inner product between the learned graph representation $z$, as in \cite{kipf2016variational}. We train each \acrshort{gae} for 50 and 150 epochs, respectively, for DTC and DBLP, and use the Adam optimiser with a learning rate of $10^{-3}$ and $10^{-4}$. We rely on L2 regularisation for the logistic regressor and use the default parameters of the scikit-learn package. We implement \acrshort{our} based on the GRETEL framework \cite{prado2023developing,prado2022gretel}. We did not perform any hyperparameter optimisation for \acrshort{our}.

\subsection{Discussion}\label{sec:discussion}

Table \ref{tab:dygrace_performance} depicts the performance of \acrshort{our}. We report averages on 10-fold cross-validation. We reserve $10\%$ of the first snapshot as test data and adapt the GAEs and the logistic regressor in an online fashion for the other snapshots. As a preliminary assessment of the performances of \acrshort{our}, we employ omniscient oracles for both datasets such that the correctness refers to the accuracy of the explainer w.r.t. the ground truth. Excluding the oracles' performances allows the reader to understand each explainer's limitations and benefits better. Where applicable, we report metrics $@1$ and $@k = 10$. As anticipated in Sec. \ref{sec:method}, \acrshort{our} accesses the oracle only in the first snapshot while relying on the \acrshort{gae}s in the successive snapshots.

In both datasets, \acrshort{our} has satisfactory results regarding correctness $@k$. Notice, however, that DBLP, being a real-world scenario, is far more complex than DTC. In DBLP, the ego networks belonging to the two classes share a similar structure, with the sole difference in the edge weights. Therefore, the correctness $@1$ in this scenario fluctuates (i.e., increasing until $t_3$, $t_6$, and decreasing afterwards). We believe this happens due to similar latent spaces that the two GAEs learn, which cannot completely distinguish between factual and counterfactual graphs. 

It is interesting to notice that the correctness $@k$ has a non-decreasing trend for DTC, meaning that valid counterfactuals might not be the most probable w.r.t. the input graph. However, they get captured by the underlying logistic regressor. Meanwhile, in DBLP, the correctness $@1$ and $@k$ degrades after iteration $t_7$, meaning that the structure of the graphs mutates heavily, making the two GAEs unable to correctly represent the two classes. The GED follows a similar trend throughout the iterations, indicating that the logistic regressor does not need to go far away from the separating hyperplane to fetch valid counterfactuals. Additionally, this phenomenon suggests that the logistic regressor "pays attention" to the first two components of Eq. \ref{eq:optimisation_func} to produce counterfactuals rather than concentrating more on the similarity of the instance with its potential counterfactual.

One drawback that could hinder \acrshort{our}'s usability is the running time\footnote{Notice that in iterations $t_8,t_9,t_{10}$ in DBLP, \acrshort{our} fails to produce counterfactuals in the first fold, thus finishing the search for valid counterfactuals prematurely. Therefore, the running time is reduced by a factor of $2$ w.r.t. the previous iterations.}, especially in successive iterations (see DBLP) where there are distributional shifts and the two GAEs need to update. However, one could implement an update trigger mechanism only in those scenarios where substantial drifts happen, which can get signalled according to a statistical test w.r.t. the reconstruction errors of the current and previous iterations (see Sec. \ref{sec:method}).

\begin{table}[!t]
\centering
\caption{Average of \acrshort{our}'s performance on DynTree-Cycles (DTC) and DBLP-Coauthors (DBLP) on 10-fold cross-validation.}
\label{tab:dygrace_performance}
\resizebox{\textwidth}{!}{%
\begin{tabular}{llccccccccccccccccccc}
\toprule
\multicolumn{4}{l}{\multirow{2}{*}{}} &
  \multicolumn{1}{c}{Runtime (s) $\downarrow$} &
   &
  \multicolumn{3}{c}{Correctness $\uparrow$} &
   &
  \multicolumn{3}{c}{Sparsity $\downarrow$} &
   &
  \multicolumn{3}{c}{GED $\downarrow$} &
   &
   &
  \multicolumn{1}{l}{Oracle Calls $\downarrow$} \\ \cmidrule{7-9} \cmidrule{11-13} \cmidrule{15-17}
\multicolumn{4}{l}{} &
  \multicolumn{1}{c}{} &
   &
  \multicolumn{1}{c}{@1} &
   &
  \multicolumn{1}{c}{@k} &
   &
  \multicolumn{1}{c}{@1} &
   &
  \multicolumn{1}{c}{@k} &
   &
  \multicolumn{1}{c}{@1} &
   &
  \multicolumn{1}{c}{@k} &
   &
   &
  \multicolumn{1}{l}{} \\ \midrule
\multirow{4}{*}{\rotatebox{90}{DTC}} &
   &
  $t_0$ &
   &
  \multicolumn{1}{c}{$24.73^{\pm70.419}$} &
   &
  \multicolumn{1}{c}{$0.40^{\pm 0.490}$} &
   &
  \multicolumn{1}{c}{$1.00^{\pm 0.000}$} &
   &
  \multicolumn{1}{c}{$0.65^{\pm 0.109}$} &
   &
  \multicolumn{1}{c}{$0.72^{\pm 0.114}$} &
   &
  \multicolumn{1}{c}{$36.90^{\pm 5.839}$} &
   &
  \multicolumn{1}{c}{$40.04^{\pm 6.272}$} &
   &
   &
   $9000.00^{\pm 0.00}$
   \\
 &
   &
  $t_1$ &
   &
  \multicolumn{1}{c}{$73.35^{\pm3.622}$} &
   &
  \multicolumn{1}{c}{$0.60^{\pm 0.490}$} &
   &
  \multicolumn{1}{c}{$1.00^{\pm 0.000}$} &
   &
  \multicolumn{1}{c}{$0.69^{\pm 0.087}$} &
   &
  \multicolumn{1}{c}{$0.73^{\pm 0.085}$} &
   &
  \multicolumn{1}{c}{$38.60^{\pm 4.652}$} &
   &
  \multicolumn{1}{c}{$40.46^{\pm 4.725}$} &
   &
   &
  $0.00^{\pm 0.00}$ \\
 &
   &
  $t_2$ &
   &
  \multicolumn{1}{c}{$74.62^{\pm6.225}$} &
   &
  \multicolumn{1}{c}{$0.70^{\pm 0.458}$} &
   &
  \multicolumn{1}{c}{$1.00^{\pm 0.000}$} &
   &
  \multicolumn{1}{c}{$0.73^{\pm 0.038}$} &
   &
  \multicolumn{1}{c}{$0.75^{\pm 0.079}$} &
   &
  \multicolumn{1}{c}{$40.50^{\pm 2.291}$} &
   &
  \multicolumn{1}{c}{$41.68^{\pm 4.366}$} &
   &
   &
  $0.00^{\pm 0.00}$ \\
 &
   &
  $t_3$ &
   &
  \multicolumn{1}{c}{$65.99^{\pm8.409}$} &
   &
  \multicolumn{1}{c}{$0.40^{\pm 0.490}$} &
   &
  \multicolumn{1}{c}{$1.00^{\pm 0.000}$} &
   &
  \multicolumn{1}{c}{$0.76^{\pm 0.065}$} &
   &
  \multicolumn{1}{c}{$0.74^{\pm 0.090}$} &
   &
  \multicolumn{1}{c}{$42.50^{\pm 3.775}$} &
   &
  \multicolumn{1}{c}{$41.16^{\pm 4.983}$} &
   &
   &
  $0.00^{\pm 0.00}$ \\ \midrule
\multirow{11}{*}{\rotatebox{90}{DBLP}} &
   &
  $t_{0}$ &
   &
  $12.83^{\pm 32.599}$ &
   &
  $0.38^{\pm 0.484}$ &
   &
  $1.00^{\pm 0.000}$ &
   &
  $0.67^{\pm 0.186}$ &
   &
  $0.71^{\pm 0.194}$ &
   &
  $28.50^{\pm 8.617}$ &
   &
  $30.04^{\pm 9.499}$ &
   &
   &
   $9900.00^{\pm 0.00}$
   \\
 &
   &
  $t_{1}$ &
   &
  $140.21^{\pm 53.821}$ &
   &
  $0.25^{\pm 0.433}$ &
   &
  $0.88^{\pm 0.331}$ &
   &
  $0.54^{\pm 0.315}$ &
   &
  $0.62^{\pm 0.238}$ &
   &
  $35.25^{\pm 47.872}$ &
   &
  $40.24^{\pm 48.264}$ &
   &
   &
  $0.000^{\pm 0.00}$ \\
 &
   &
  $t_{2}$ &
   &
  $121.29^{\pm 57.151}$ &
   &
  $0.12^{\pm 0.331}$ &
   &
  $0.88^{\pm 0.331}$ &
   &
  $1.52^{\pm 2.709}$ &
   &
  $3.59^{\pm 6.178}$ &
   &
  $18.88^{\pm 15.070}$ &
   &
  $34.82^{\pm 33.389}$ &
   &
   &
  $0.000^{\pm 0.00}$ \\
 &
   &
  $t_{3}$ &
   &
  $150.73^{\pm 20.020}$ &
   &
  $0.38^{\pm 0.484}$ &
   &
  $0.88^{\pm 0.331}$ &
   &
  $0.85^{\pm 0.489}$ &
   &
  $0.81^{\pm 0.384}$ &
   &
  $63.12^{\pm 51.033}$ &
   &
  $57.86^{\pm 44.450}$ &
   &
   &
  $0.000^{\pm 0.00}$ \\
 &
   &
  $t_{4}$ &
   &
  $161.89^{\pm 10.609}$ &
   &
  $0.25^{\pm 0.433}$ &
   &
  $0.62^{\pm 0.484}$ &
   &
  $0.63^{\pm 0.261}$ &
   &
  $0.58^{\pm 0.263}$ &
   &
  $21.62^{\pm 15.337}$ &
   &
  $20.51^{\pm 15.186}$ &
   &
   &
  $0.000^{\pm 0.00}$ \\
 &
   &
  $t_{5}$ &
   &
  $162.77^{\pm 13.782}$ &
   &
  $0.50^{\pm 0.500}$ &
   &
  $1.00^{\pm 0.000}$ &
   &
  $1.01^{\pm 1.388}$ &
   &
  $0.75^{\pm 0.835}$ &
   &
  $21.12^{\pm 10.361}$ &
   &
  $21.71^{\pm 12.553}$ &
   &
   &
  $0.000^{\pm 0.00}$ \\
 &
   &
  $t_{6}$ &
   &
  $165.39^{\pm 15.160}$ &
   &
  $0.50^{\pm 0.500}$ &
   &
  $1.00^{\pm 0.000}$ &
   &
  $2.26^{\pm 3.125}$ &
   &
  $2.21^{\pm 4.238}$ &
   &
  $55.00^{\pm 39.528}$ &
   &
  $53.36^{\pm 42.852}$ &
   &
   &
  $0.000^{\pm 0.00}$ \\
 &
   &
  $t_{7}$ &
   &
  $81.22^{\pm 82.797}$ &
   &
  $0.12^{\pm 0.331}$ &
   &
  $0.50^{\pm 0.500}$ &
   &
  $1.44^{\pm 3.244}$ &
   &
  $3.13^{\pm 5.679}$ &
   &
  $11.62^{\pm 19.118}$ &
   &
  $23.80^{\pm 32.100}$ &
   &
   &
  $0.000^{\pm 0.00}$ \\
 &
   &
  $t_{8}$ &
   &
  $76.85^{\pm 75.487}$ &
   &
  $0.12^{\pm 0.331}$ &
   &
  $0.50^{\pm 0.500}$ &
   &
  $0.45^{\pm 0.535}$ &
   &
  $0.84^{\pm 0.671}$ &
   &
  $17.88^{\pm 23.861}$ &
   &
  $33.32^{\pm 24.627}$ &
   &
   &
  $0.000^{\pm 0.00}$ \\
 &
   &
  $t_{9}$ &
   &
  $66.97^{\pm 67.959}$ &
   &
  $0.25^{\pm 0.433}$ &
   &
  $0.50^{\pm 0.500}$ &
   &
  $0.44^{\pm 0.450}$ &
   &
  $0.67^{\pm 0.257}$ &
   &
  $33.75^{\pm 34.387}$ &
   &
  $51.85^{\pm 20.576}$ &
   &
   &
  $0.000^{\pm 0.00}$ \\
 &
   &
  $t_{10}$ &
   &
  $63.07^{\pm 65.097}$ &
   &
  $0.25^{\pm 0.433}$ &
   &
  $0.38^{\pm 0.484}$ &
   &
  $0.90^{\pm 1.310}$ &
   &
  $0.97^{\pm 0.685}$ &
   &
  $57.62^{\pm 69.611}$ &
   &
  $77.09^{\pm 62.334}$ &
   &
   &
  $0.000^{\pm 0.00}$ \\ \bottomrule
\end{tabular}%
}
\end{table}

\section{Conclusion}\label{sec:conclusion}

We demonstrated the effectiveness of our semi-supervised Graph Counterfactual Explainer across synthetic and real-world datasets. Deployed on the synthetic Tree-Cycles dataset and the real-world DBLP-Coauthors dataset, \acrshort{our} showcased satisfactory results in terms of correctness. Despite the complexity of the DBLP dataset due to its real-world nature, \acrshort{our} managed to discern between factual and counterfactual graphs to a large extent. 

The continual trend of increasing correctness in both datasets asserts \acrshort{our}'s ability to capture valid counterfactuals, even when they may not be the most probable concerning the input graph. This is particularly impressive given that GED remained stable throughout iterations, indicating that the model doesn't have to deviate significantly from the separating hyperplane to identify valid counterfactuals. Nonetheless, \acrshort{our}'s potential downside lies in its extensive runtime during consecutive iterations, particularly visible in datasets with significant distributional shifts. We suggest implementing an update trigger mechanism that activates only when substantial drifts occur to alleviate this issue. This approach would rely on a statistical test concerning the reconstruction errors of current and previous iterations.

Our findings support \acrshort{our} as a promising, flexible framework that can learn meaningful graph representations for counterfactual explanations. Thus, \acrshort{our} offers exciting new opportunities for future research.

\bibliographystyle{unsrtnat}
\bibliography{references}  






\end{document}